\pdfoutput=1
\documentclass{article}
\PassOptionsToPackage{hyphens}{url}

\usepackage{microtype}
\usepackage{graphicx}
\usepackage{booktabs}
\usepackage[dvipsnames]{xcolor}
\usepackage{amsmath,amssymb}
\usepackage{mathtools}
\usepackage{enumitem}
\usepackage{needspace}

\usepackage{hyperref}
\hypersetup{colorlinks=true,linkcolor=teal!60!black,citecolor=teal!60!black,urlcolor=teal!60!black}

\usepackage[accepted]{icml2026}
\makeatletter
\renewcommand{\Notice@String}{Preprint. Preliminary results, shared for discussion; not peer reviewed.}
\makeatother

\newcommand{\disj}{\textsc{disjoint}}
\newcommand{\trainfwd}{\texttt{train-fwd}}

\icmltitlerunning{DreamingGoose: Staged Distillation to Recurrent Diffusion LMs}

\begin{document}

\twocolumn[
\icmltitle{DreamingGoose: Staged Distillation from Autoregressive Transformers\\
to Bidirectional Recurrent Diffusion Language Models}

\icmlsetsymbol{equal}{*}

\begin{icmlauthorlist}
\icmlauthor{Julian Boesch}{purdue,obit}
\icmlauthor{Andrew Wee}{purdue,obit}
\icmlauthor{Alexander Stranzl}{stonybrook}
\end{icmlauthorlist}

\icmlaffiliation{purdue}{Purdue University}
\icmlaffiliation{obit}{Obit Research}
\icmlaffiliation{stonybrook}{State University of New York at Stony Brook}

\icmlcorrespondingauthor{Julian Boesch}{jboesch@obitmc.com}
\icmlcorrespondingauthor{Andrew Wee}{awee@obitmc.com}
\icmlcorrespondingauthor{Alexander Stranzl}{alexander.stranzl@stonybrook.edu}

\icmlkeywords{diffusion language models, linear attention, distillation, model conversion, curriculum learning, associative recall, Machine Learning, ICML}

\vskip 0.3in
]

\printAffiliationsAndNotice{}

\makeatletter
\gdef\@icmltitlerunning{DreamingGoose: Staged Distillation to Recurrent Diffusion LMs}
\makeatother

\begin{abstract}
Pretrained autoregressive Transformers represent a large sunk investment in compute. Existing conversion methods reuse that investment by changing either the architecture (attention to recurrence) or the objective (next-token prediction to denoising), never both. We convert Qwen3 teachers at $1.7$B and $8$B into attention-free, bidirectional, gated-delta-rule diffusion students in three stages, so that each capability can be traced to the stage that kept or lost it. Language modeling transfers only partially and in-distribution; in-context retrieval does not transfer. On a multi-query recall probe where the teachers score $0.34$--$0.58$, both converted students score $0.000$, and diffusion pretraining alone does not restore retrieval. A retrieval curriculum in the final stage, which gradually lengthens the gap between a key--value table and the queries that address it, restores it only stochastically: on a fixed schedule, one seed in three learns to retrieve. Advancing the gap only while a running accuracy estimate stays above a threshold works for all three of those seeds, holds on real text, and carries unchanged to $8$B, where two of three seeds succeed. The third had not learned within its fixed 16k-step budget: retrieval switches on abruptly at a seed-dependent step ($6.5$k and $11$k in the other two), so a fixed budget can cut a late run off. One boundary survives every intervention: every model that learns retrieval scores $0.000$ on tokens that never appeared in a retrieval episode, and an arm that resamples the key and value tokens every batch shows this is a coverage limit, not memorization of particular bindings. Separately, we convert a $7$B code model into a $3{:}1$ recurrent--attention block-diffusion hybrid over $85$k steps and report two negative training results.
\end{abstract}

\section{Introduction}\label{sec:intro}

Two conversion literatures exist, one for each of two axes, and they do not intersect. Along the objective axis, pretrained autoregressive Transformers are adapted into \emph{diffusion} language models, with the Transformer backbone kept \citep{gong2025diffullama,ye2025dream}. Along the architecture axis, attention is distilled into \emph{linear-time recurrence}, with next-token prediction kept \citep{goldstein2025radlads}. An attention-free diffusion LM built from a pretrained Transformer has to move along both axes; to our knowledge, no prior work has done both.

Our conversion does both, in three stages, each with its own objective and its own set of frozen parameters (\S\ref{sec:pipeline}): distill the attention teacher into a causal recurrent student; expand that student into a bidirectional denoiser; continue pretraining under a masked-diffusion objective. The staging is the method: when a capability fails to transfer, we can say at which stage it was lost.

The finding that organizes the paper is a dissociation. Converted students keep part of what the teacher \emph{knows}, and only in-distribution, but lose all of what the teacher can \emph{do within a sequence}. On a multi-query associative recall probe (in its causal form, \S\ref{sec:probe}) where the teachers score $0.335$--$0.575$, the converted students score $0.000$ at both scales: retrieval is not degraded but absent, and diffusion pretraining alone does not bring it back. Our companion paper \citep{anatomyrecall} reproduces this failure in a from-scratch recurrent model on a synthetic recall task, and removes it with a curriculum that slowly lengthens the gap between a key--value table and the queries that address it. We port that curriculum into the final stage.

The port works, but stochastically; making it reliable is the paper's main practical result. Under an \emph{open-loop} curriculum, with the gap advanced on a fixed schedule, one seed in three learns to retrieve. Advancing the gap only while measured retrieval accuracy stays above a threshold, a \emph{closed-loop} schedule, brings the same three seeds to $3/3$. The method then carries unchanged to $8$B, where two of three seeds succeed (\S\ref{sec:8b}).

\needspace{3\baselineskip}
\textbf{Contributions.}
\begin{enumerate}[leftmargin=1.4em,itemsep=2pt]
\item \textbf{A staged conversion pipeline across both axes} (\S\ref{sec:pipeline}) and, because it is staged, an account of what transfers and where it is lost: language modeling transfers partially and only in-distribution; in-context retrieval does not transfer at all ($0.000$ at $1.7$B and $8$B; \S\ref{sec:transfer}).
\item \textbf{A closed-loop retrieval curriculum} (\S\ref{sec:stage3}, \S\ref{sec:lottery}). With the gap advanced on a fixed schedule, the curriculum repairs retrieval in one seed of three. A closed-loop schedule lifts the same seeds to $3/3$ (per-seed means $0.97$, $0.98$, $0.97$) and holds on real text ($0.955$--$0.995$ on every evaluated grid cell).
\item \textbf{Transfer to 8B, and the residual failure mode} (\S\ref{sec:8b}). The best seed's grid mean rises from $0.000$ to $0.90$ and generalizes to twice the trained pair count and beyond the trained gap. Two of three seeds succeed. Retrieval switches on at a step that varies widely by seed ($6.5$k, $11$k, ${>}16$k steps), and the third's miss is consistent with a fixed step budget cutting off a late onset.
\item \textbf{A token-coverage boundary} (\S\ref{sec:coverage}). Every model that learns to retrieve over its training tokens scores $0.000$ on held-out tokens, at both scales. An arm that resamples the key and value tokens every batch rules out memorized bindings as the cause; three ways of pacing the curriculum over randomized pools fail through three different mechanisms, two of them failures of the pacing policy and one consistent with too short a budget.
\item \textbf{A block-diffusion conversion of a $7$B code model} into a $3{:}1$ recurrent--attention hybrid (\S\ref{sec:blockdiff}). The final checkpoint reaches held-out block-diffusion NLL $2.819$ after two rollbacks and a learning-rate reduction. We report two negative results from training it: an auxiliary representation-alignment loss that brings no benefit, and a learning-rate ceiling that appears only after 50k stable steps.
\end{enumerate}

\textbf{Terminology.} An \emph{arm} is a training configuration; a \emph{run} is one seed of an arm. A \emph{band} is a fixed set of vocabulary tokens used as keys and values; a model is evaluated both on the band it trained on and on a held-out one. A \emph{pool} is the set of tokens from which a randomized-pool arm draws fresh keys and values every batch. The \emph{gap} is the number of haystack tokens between a key--value table and the queries that address it. Learning to retrieve is a binary event in the training telemetry: the exponential moving average (EMA) of retrieval accuracy sits near zero for thousands of steps, then rises past the curriculum's gate threshold (\S\ref{sec:stage3}) within roughly a thousand. We call that rise the \emph{avalanche}, the step at which it begins the \emph{onset}, and a run that avalanches within its budget \emph{locked in}. Because the event is binary, we report per-seed outcomes and lock-in counts rather than accuracy averaged over seeds. \emph{Conversion scale} refers to experiments on real pretrained models, as opposed to the synthetic recall task of \citet{anatomyrecall}.

\section{Related Work}\label{sec:related}

\textbf{AR-to-diffusion adaptation.} \citet{gong2025diffullama} and \citet{ye2025dream} adapt pretrained autoregressive Transformers into masked diffusion LMs, establishing that the objective change is survivable and far cheaper than pretraining. Both retain attention.

\textbf{Attention-to-recurrence distillation.} RADLADS \citep{goldstein2025radlads} converts attention into RWKV-7-style recurrence at multi-billion scale on a few hundred million tokens, retaining next-token prediction; our stage 1 follows its two-phase recipe.

\textbf{Recurrent denoisers.} DiffuMamba \citep{singh2025diffumamba} shows that scan backbones are viable denoisers. We convert into cells with the rank-1 state update of \citet{yang2024gateddeltanet} and \citet{peng2025rwkv7}: the gated delta rule by default, and RWKV-7 in one $8$B arm (\S\ref{sec:8b}).

\textbf{Block diffusion} \citep{arriola2025bd3lm} interpolates between the autoregressive and diffusion objectives; it is the objective of our conversion in \S\ref{sec:blockdiff}.

\textbf{Recall in recurrence.} It is well documented that fixed-state recurrences underperform attention on associative recall \citep{arora2023zoology,arora2024based}. Our companion paper argues on a synthetic bench that the deficit is substantially a \emph{learnability} problem rather than a capacity limit \citep{anatomyrecall}, and convergent evidence appears elsewhere \citep{okpekpe2025recall,blouir2024birdie}; this paper tests that reading at scale.

\section{Method: The Conversion Pipeline}\label{sec:pipeline}

The pipeline takes an attention teacher to an attention-free bidirectional recurrent diffusion student in three stages. Table~\ref{tab:settings} lists the concrete settings; the final stage is configured identically at both scales apart from the corpus, the retrieval-episode probability and, for the one RWKV-7 arm, the learning rate.

\subsection{Stage 1: attention to recurrence}\label{sec:stage1}

This stage changes the architecture and nothing else: attention becomes recurrence, and the objective stays next-token prediction.

The teacher is a standard Transformer in which every token-mixing sublayer is softmax attention: Qwen3-1.7B (28 layers, hidden size $2048$) or Qwen3-8B (36 layers, hidden size $4096$, 32 query and 8 key--value heads) \citep{yang2025qwen3}. The student inherits the teacher's token embeddings, MLPs, normalization layers and output head, and replaces \emph{every} attention sublayer with a gated delta-rule recurrent cell \citep{yang2024gateddeltanet}; the new cells are initialized randomly, and everything inherited stays frozen.

Training follows RADLADS \citep{goldstein2025radlads} in two phases. \emph{Attention transfer} trains each cell in isolation, on the teacher's own hidden states, to reproduce the output of the attention sublayer it replaces. \emph{Refinement} then trains the assembled student end-to-end to minimize the KL divergence between its next-token distribution and the teacher's.

At $8$B the teacher runs in 8-bit precision. The corpus is a 12M-character slice of text8 \citep{mahoney2011text8}, lowercase Wikipedia text (2.36M unique tokens), at sequence length 256. The refinement budget is 60k steps at batch 2 (30.7M tokens) at $1.7$B and 120k steps at batch 1 (${\approx}30$M tokens) at $8$B, about 13 passes over the slice. We report stage-1 quality as \emph{top-1 agreement} (the fraction of positions where the student's and teacher's argmax tokens coincide) and \emph{perplexity retention} (teacher perplexity divided by student perplexity, so $1.0$ is parity). On the distillation text itself, agreement is $0.746$ at $1.7$B and $0.779$ at $8$B, and retention is $0.932$ at $1.7$B; these are in-sample numbers. On held-out text8 at $8$B, agreement falls to $0.544$ and retention to $0.435$, and off-domain retention collapses ($0.056$ on English prose, $0.0011$ on code). We did not measure held-out quality at $1.7$B.

\subsection{Stage 2: bidirectional expansion}\label{sec:stage2}

This stage is structural: it adds parameters and trains none of them, so the new capacity is learned in stage 3. Each converted layer gains a \emph{backward} cell, initialized as a copy of its forward cell and run over the reversed sequence, plus a merge matrix $W \in \mathbb{R}^{d \times 2d}$ that combines the two streams, initialized to their average. The embedding table gains one row for a mask token. Bidirectionality is what a denoiser requires: every position must see both sides of the sequence. It also lets the student read each query before the table the query must search, because the backward stream reaches the query before the text that precedes it.

\subsection{Stage 3: diffusion continual pretraining, with a retrieval curriculum}\label{sec:stage3}

\textbf{Objective.} Masked discrete diffusion \citep{austin2021d3pm,sahoo2024mdlm}: a fraction of tokens is replaced by the mask token and the model is trained to recover them, with the mask ratio annealed from $0.15$ to $0.65$ over training. By default the trainable set is the backward cells, the merges and the mask embedding, with the forward cells frozen at their stage-1 values. The \trainfwd{} variant also unfreezes the forward cells. Every curriculum result in \S\ref{sec:results} uses \trainfwd{}; \S\ref{sec:transfer} shows that the default, with the forward cells frozen, did not learn to bind in the one pilot run that tested it.

\textbf{Retrieval episodes.} With probability $p$ a training batch is a retrieval episode rather than a text batch. An episode has the layout
\begin{center}\small
$[\text{text prefix}]\;[k_1\,v_1\;\cdots\;k_N\,v_N]\;[\text{text haystack, } g \text{ tokens}]$\\[2pt]
$[k_j\;\texttt{MASK}]\;[k_{j'}\;\texttt{MASK}]\;\cdots$
\end{center}
with $N \in \{4, 8, 16\}$ pairs and four queried pairs per episode, every answer masked and supervised. Keys and values are single frequent tokens of the corpus, 64 of each, fixed for the run; the prefix and haystack are real text. The haystack length $g$ (the table-to-query distance, or \emph{gap}) is the curriculum variable: each batch samples $g$ uniformly in $[0, \text{cap}]$, and the cap grows from $0$ toward the maximum that fits the sequence. Evaluation spans the full gap grid regardless of the current cap.

\textbf{Schedule: open-loop vs.\ closed-loop.} In the \emph{open-loop} schedule the cap grows linearly with the step count, reaching its maximum at 80\% of training. The \emph{closed-loop} schedule tracks the EMA of retrieval accuracy over retrieval batches, and the cap advances by a fixed increment only on steps where the EMA is at or above a threshold ($0.6$). In this schedule the cap is a ratchet: it never retreats. A \emph{thermostat} variant, used in \S\ref{sec:coverage}, lets the cap retreat while the EMA is below its threshold (threshold $0.5$; the cap retreats three increments per failing step). We call the ratchet at $0.6$, the ratchet at $0.25$ and the thermostat the three \emph{gate policies} of \S\ref{sec:coverage}.

\textbf{Randomized symbol pool.} For the coverage experiments of \S\ref{sec:coverage}, keys and values are drawn afresh for every batch from one shared pool of $P$ tokens ($P \in \{128, 512\}$, excluding the held-out evaluation band), rather than from the fixed 64 keys and 64 values, so that no particular binding can be memorized. Pool-size \emph{annealing} moves up through tiers ($128 \to 512 \to 2048$), advancing after a stretch of consecutive steps with the EMA above threshold.

\subsection{Evaluation: the recall probe, its controls, and a teacher check}\label{sec:probe}

Retrieval is measured with a multi-query associative recall probe in the episode format above: $N \in \{4, 8, 16\}$ pairs, a real-text haystack of $g \in \{0, 64, 160\}$ tokens, and one masked answer position scored per episode by exact match. Chance is the value marginal, ${\approx}0.023$: the accuracy of guessing values by their corpus frequency. Exact match scores $0.000$ for a model that never emits a value token at the answer position, so a failed model can score either $0.000$ or about $0.02$ depending on how it fails; both mean no retrieval, and we report observed values throughout. Two controls accompany every number. \emph{Shuffled}: the bindings are permuted and the original value is scored, so a genuine retriever's score must drop sharply. \emph{Mismatch}: the query is a key absent from the table, which measures what the model emits when there is no binding to retrieve; we call this the \emph{mismatch baseline}. The teachers and the stage-1 students are causal, so they are scored on a causal form of the probe: the table and one query key, no haystack, and the value scored as the next token (marginal $0.016$). We validated this format on both teachers before scoring the students: a teacher must clear the marginal by a wide margin and drop sharply under shuffling. This \emph{teacher check} rules out probe artifacts as an explanation for student failures.

\textbf{Trained band and disjoint band.} The \disj{} evaluation is the identical probe with keys and values drawn from a band that no retrieval episode and no randomized pool ever touches during training. It asks whether the model learned retrieval in general or retrieval over particular tokens.

\begin{table}[t]
\centering
\caption{\textbf{Settings at both scales.} Stage 3 uses batch 4, sequence length 256, learning rate $5{\times}10^{-4}$ (the RWKV-7 arm: $1{\times}10^{-4}$, with gradient checkpointing) and a standard 16k-step budget at both scales (one 8B arm in \S\ref{sec:coverage} used 24k). In-sample agreement is measured on the distillation text (\S\ref{sec:stage1}). Cells with two values give the two $1.7$B arms (text8, wikitext); the $8$B run uses text8 with $p{=}0.5$. $p$ is the retrieval-episode probability. Wall times are on different hardware and are not comparable.}
\label{tab:settings}
\vskip 0.1in
\scriptsize
\setlength{\tabcolsep}{2pt}
\begin{tabular}{@{}p{0.30\columnwidth}p{0.32\columnwidth}p{0.32\columnwidth}@{}}
\toprule
 & $1.7$B & $8$B \\
\midrule
teacher & Qwen3-1.7B & Qwen3-8B \\
student cells & gated delta rule & gated delta rule; RWKV-7 in one arm \\
stage-1 refinement & 60k steps, batch 2 (30.7M tokens) & 120k steps, batch 1 (${\approx}$30M tokens) \\
stage-1 agreement, in-sample & $0.746$ & $0.779$ \\
stage-1 agreement, held-out & not measured & $0.544$ \\
stage-3 corpus & text8 / wikitext & text8 \\
$p$ (retrieval) & $0.7$ / $0.5$ & $0.5$ \\
gate threshold & $0.6$ & $0.6$ \\
hardware & RTX 3090, 24 GB & RTX PRO 6000, 96 GB \\
wall time per arm & ${\approx}6$ h & ${\approx}2.6$ h \\
\bottomrule
\end{tabular}
\end{table}

\section{Results}\label{sec:results}

\S\ref{sec:transfer} reports what stage 1 transfers and what the curriculum requires. \S\ref{sec:lottery} shows that the open-loop curriculum is a lottery at $1.7$B and that closing the loop fixes it. \S\ref{sec:8b} applies the recipe unchanged at $8$B. \S\ref{sec:coverage} reports the one boundary that no intervention moved.

\subsection{What transfers, and what does not}\label{sec:transfer}

Stage 1 transfers language modeling only partially: top-1 agreement with the teacher is $0.746$ at $1.7$B and $0.779$ at $8$B on the distillation text, but $0.544$ on held-out text at $8$B, and perplexity retention collapses off-domain (\S\ref{sec:stage1}). It does not transfer in-context retrieval at all. On the causal recall probe (\S\ref{sec:probe}) the teachers score $0.335$--$0.350$ at $1.7$B and $0.385$--$0.575$ at $8$B for $N \in \{4, 8, 16\}$, and drop to $0.025$--$0.105$ when the bindings are shuffled. The converted students score $0.000$ at both scales, and their accuracy is insensitive to whether the bindings are shuffled: they emit the mismatch baseline rather than retrieve. Under stage-3 diffusion pretraining alone (no retrieval episodes), retrieval stays at $0.000$. The capacity exists: the same cell trained from scratch solves 16-pair recall at $0.996$ on the synthetic bench \citep{anatomyrecall}. The distillation did not instill it.

\textbf{The curriculum needs trainable forward cells.} In a single pilot run, injecting retrieval episodes into stage 3 with the forward cells frozen taught the model to emit well-formed answers but not to bind: accuracy stayed at the mismatch baseline. With the forward cells unfrozen (\trainfwd), the model acquires retrieval. This is what the companion paper's mechanism predicts: if the circuit is \emph{mark-and-route}---the backward stream marks the binding as it passes the query, and the forward stream routes it to the answer position---then the routing direction must be trainable. The cost is a small denoising penalty relative to the no-curriculum control ($-0.06$ masked-token accuracy, $0.144$ against $0.204$, in the same pilot).

\subsection{The lottery, and the closed-loop fix}\label{sec:lottery}

\begin{table}[t]
\centering
\caption{\textbf{Conversion-scale replication.} Retrieval on the trained token band, per seed (mean over the $N\in\{4,8,16\}\times\text{gap}\in\{0,64,160\}$ grid; marginal ${\approx}0.023$). The wikitext row is a single run; its cell gives the grid-cell range rather than a per-seed mean. \disj{} is the identical probe with keys and values from a held-out token band. The open-loop row's non-zero seed is a partial lock-in ($0.40$--$0.70$ across grid cells); the closed-loop schedule lifts the same three seeds to ceiling. Every row---including every solved one---reads $0.000$ on the disjoint band.}
\label{tab:atscale}
\vskip 0.1in
\scriptsize
\setlength{\tabcolsep}{3.5pt}
\begin{tabular}{@{}llcc@{}}
\toprule
setting & lock-in & per-seed (trained) & disj. \\
\midrule
$1.7$B, open-loop          & 1/3          & $0.00$/$0.00$/$0.55$ & $0.000$ \\
$1.7$B, \textbf{closed-loop} & \textbf{3/3} & $\mathbf{0.97}$/$\mathbf{0.98}$/$\mathbf{0.97}$ & $0.000$ \\
$1.7$B, closed-loop, wikitext & 1/1       & $0.955$--$0.995$ (range) & $0.000$ \\
$8$B, closed-loop          & 2/3          & $0.90$/$0.78$/chance & $0.000$ \\
\bottomrule
\end{tabular}
\end{table}

Table~\ref{tab:atscale} collects the conversion-scale results, and the pattern is plain: the open-loop curriculum is a lottery, and on the same three seeds the closed-loop one is not. Under the open-loop schedule at $1.7$B, three seeds draw $0.00$, $0.00$ and $0.55$, the last ranging over $0.40$--$0.70$ across grid cells: one partial lock-in in three. An earlier unseeded pilot run of the same configuration had reached $0.94$--$1.00$; an exact rerun returned $0.000$, and we count the pilot as one more draw from the same lottery rather than as a result. The synthetic bench taught the same lesson independently: a difference that held at two seeds reversed sign at ten \citep{anatomyrecall}. In this regime, a mean over a few seeds is not evidence.

\textbf{Why the open-loop schedule fails.} The cap advances on a clock. A run that has not yet learned to bind at short gaps is dragged to long ones regardless, and spends the rest of its budget training at distances it cannot solve. Whether a seed reaches onset before the cap outruns it is close to a coin flip.

\textbf{The closed-loop schedule.} Gating the cap on the retrieval EMA removes the failure: the cap never advances while a run is failing, and a run that stalls simply trains longer at its current difficulty. On the same three seeds, all three lock in under the closed-loop schedule (per-seed means $0.97$, $0.98$, $0.97$). Accuracy at the maximum evaluated gap matches accuracy at gap zero, and both controls---shuffled and mismatch---are clean, as they are for every locked-in run. The telemetry agrees: all three advanced the cap with the retrieval EMA at $0.98$--$0.99$. None reached the maximum cap (they stopped at $0.66$--$0.87$ of it), yet all are at ceiling at the maximum evaluated gap, so the top 13\% of the cap range is unnecessary at this sequence length.

\textbf{Real text.} On wikitext \citep{merity2016pointer} the closed-loop recipe reaches $0.955$--$0.995$ on every $N \times \text{gap}$ cell, including $N{=}16$ at gap $160$ ($0.975$), with clean controls. Its telemetry shows why an open-loop schedule fails here: the model is still near chance at 3k steps (EMA $0.04$), so a schedule that advances the cap on a clock runs ahead of it. Onset arrives by $5.4$k steps and the cap then advances cleanly to $0.91$. Real text delays lock-in; it does not prevent it. The recipe at conversion scale is therefore \emph{curriculum $+$ trainable forward cells $+$ closed-loop schedule}, and the right summary statistic for it is a lock-in rate over seeds, not a mean accuracy.

\subsection{Scaling to 8B}\label{sec:8b}

Applied unchanged to the $8$B student, the recipe lifts the two locked-in seeds from $0.000$ to grid means of $0.90$ and $0.78$ (grid-cell range $0.69$--$0.99$ for the stronger). On denoising, the bidirectional pass beats the same model's causal pass by $0.117$ in masked-token accuracy ($0.141$ against $0.025$). Evaluated beyond its training distribution, the stronger seed still retrieves at twice the trained pair count and past the trained gap ($N \le 32$, gap $\le 224$: $0.49$--$0.99$, mean $0.79$). Both pair count and gap generalize.

\begin{figure}[t]
\centering
\includegraphics[width=0.86\columnwidth]{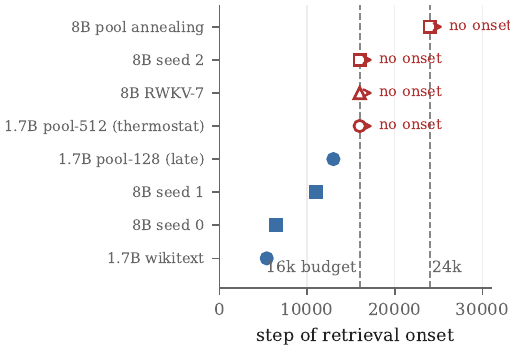}
\caption{\textbf{Onset varies widely by run, and a fixed budget truncates late onsets.} Each point is the onset step of one run, labeled by scale and run as in the text; open markers with arrows are runs that had not reached onset when their budget ended, drawn at that budget (dashed lines). Nothing is interpolated.}
\label{fig:onset}
\end{figure}

\textbf{The residual lottery appears to be onset timing.} Across three $8$B seeds the tally is $2/3$ (Figure~\ref{fig:onset}). Every locked-in run, at both scales, shows the same profile: a long dead phase, an avalanche in which retrieval NLL collapses within about a thousand steps, then a clean advance through the cap. What varies is \emph{when}. At $8$B, two seeds reached onset at $6.5$k and $11$k steps and the third had not by $16$k; at $1.7$B, onset came at $13$k steps for the randomized-pool arm of \S\ref{sec:coverage} and at $5.4$k for the wikitext arm. Onset spreads widely, and every run here had a fixed 16k-step budget, which cuts off any run whose onset comes later; with this few runs we cannot characterize the tail of the distribution. Onset also shapes final accuracy: seed 1 avalanched at $11$k and finished with a lower worst-cell accuracy ($0.35$) than seed 0, which avalanched at $6.5$k ($0.69$), because later onset leaves less training after the avalanche. The fix this points to is a closed-loop \emph{budget}---train until the avalanche, then a fixed number of steps beyond it---rather than a larger fixed total. It is designed but not yet run, so we report $2/3$ as it stands.

\textbf{A second cell family.} An RWKV-7 student at $8$B, trained with the same curriculum and schedule but at learning rate $1{\times}10^{-4}$ with coupled adapters and gradient checkpointing, trains stably and shows a smaller but positive bidirectional accuracy advantage ($+0.053$ against $+0.117$), yet does not avalanche within $16$k steps. Given the onset distribution above, we read this as underpowered rather than negative: RWKV-7's onset comes later than the gated delta-rule cell's everywhere we have measured it, and a verdict needs the closed-loop budget.

\subsection{The token-coverage boundary}\label{sec:coverage}

Every model that solves the trained-band evaluation---at $1.7$B and $8$B, including all three closed-loop seeds and the wikitext arm---scores $0.000$ on the \disj{} band. What the curriculum installs is a binding circuit over the symbols that appeared in retrieval episodes, not an abstract retrieval faculty.

The obvious explanation is memorization of fixed bindings, and it is wrong. An arm trained with keys and values drawn afresh from a $128$-token pool every batch cannot memorize any fixed binding. It locks in anyway---late, at ${\approx}13$k steps, with the same avalanche profile---and reaches $0.39$--$0.92$ on its trained band, yet it still reads $0.000$ on the disjoint band. It also generalizes in distance far beyond its own training: its cap reached only ${\approx}40$ tokens, yet gap-$64$ accuracy matches gap-$0$ accuracy, and gap-$160$ accuracy sits well above chance. The circuit is neither memorizing pairs nor distance-bound. What separates the two bands is which \emph{embeddings} ever participated in a retrieval episode.

\textbf{Three failure modes.} Randomizing the pool interacts with the gate, and we had to map that interaction before we could tell a policy failure from a refutation of the curriculum. With the ratchet at $0.6$, the pool task's EMA starts below the threshold, so the cap never advances and the arm trains only at gap ${\approx}0$ (\emph{lockout}). With the ratchet at $0.25$, the cap advances on a weak signal and the EMA then collapses; because the ratchet cannot retreat, all later training happens at gaps the model cannot solve, and the circuit never recovers (\emph{learn-then-collapse}). With the thermostat, a $128$-token pool reaches onset at $13$k steps (the arm of the previous paragraph) while a $512$-token pool never does within budget (\emph{no onset}). Two of these are failures of the pacing policy (lockout and learn-then-collapse); the third is consistent with a budget too short for a task whose onset comes much later.

\textbf{Pool annealing is untested.} The diagnosis points to pool-size annealing: reach the avalanche on a narrow pool, then widen coverage. At $8$B the one annealing run never reached onset on its first tier within a $24$k-step budget (longer than the standard $16$k), so the pool never widened. Annealing is therefore untested rather than refuted, and it is blocked on the same closed-loop budget that \S\ref{sec:8b} identifies.

\textbf{A hybrid alternative.} Because token coverage resisted every training-side intervention we tried, it gives a third reason to retain a sparse set of attention layers, beyond the usual two of state capacity and length extrapolation. An attention-retaining variant of the pipeline passes end-to-end validation on a small teacher (keeping three of its 30 attention layers), and \S\ref{sec:blockdiff} uses it at $7$B, keeping every fourth layer; its retrieval has not yet been measured.

\section{A Block-Diffusion Hybrid at 7B}\label{sec:blockdiff}

We also converted a Qwen2.5-Coder-7B-Instruct teacher \citep{hui2024qwen25coder} under a block-diffusion objective \citep{arriola2025bd3lm}. Block diffusion denoises within fixed-size blocks conditioned on committed earlier blocks; it interpolates toward autoregression and suits an instruction-following target better than full-sequence masking. This is a separate conversion: it reuses the stage-1 recipe of \S\ref{sec:stage1}, with a different initialization, and shares nothing else---a different teacher family, objective, hardware and evaluation. Unlike the students above it retains attention, for two reasons: the coverage boundary of \S\ref{sec:coverage}, and decode-time block conditioning. Pretraining ran $85{,}449$ steps ($700$M tokens) and was followed by $4{,}000$ response-masked SFT steps; reaching the final checkpoint took two rollbacks and a learning-rate reduction (Negative result 2).

\textbf{Outcome.} The result is a block-diffusion LM with a measured language-modeling baseline and a real shortfall against its teacher. We report the checkpoint at the end of the $4{,}000$-step supervised phase (step $89{,}449$). Its held-out block-diffusion NLL is $2.819$, while the EMA of the training objective ended at $2.786$, a generalization difference of $0.03$ nats. On an ordinary causal pass over held-out corpus rows, the student's NLL is $2.49$ against the teacher's $1.07$, a ${\approx}4.1\times$ perplexity ratio: the hybrid is still a noticeably worse language model than its teacher. Its free-running generation opens coherently and then collapses into repetition. The retrieval curriculum was not part of this conversion; a longer supervised phase, generation quality and decoding for this model are the subject of a forthcoming results paper.

\textbf{Architecture.} The construction differs from the pipeline above in initialization, in the absence of stage 2, and in the two-stream objective. The teacher has 28 layers (hidden size $3584$; 28 query and 4 key--value heads). The student replaces 21 of its attention layers with gated delta-rule cells and keeps every fourth one, a $3{:}1$ recurrent--attention hybrid. The cells are teacher-matched: 28 heads of key dimension $128$ and value dimension $256$, with the query and output projections copied directly from the replaced attention layer and the key/value projections tiled from its grouped-query heads. The kept attention layers are initialized as exact copies of the teacher's and remain trainable. The student inherits embeddings, MLPs, norms and the output head; among these only the embeddings are trainable. Trainable parameters total $2.91$B.

Stage 1 is as in \S\ref{sec:stage1}, except that the copied projections replace random initialization and refinement stops on an agreement plateau. Stage 2 is not needed: block diffusion is bidirectional only within a block, and the within-block context is provided by the noised stream described next rather than by a backward recurrence.

\textbf{Two-stream training.} Each step builds a clean and a noised view of the same batch. The clean stream is an ordinary causal pass of the network. Within each 64-token block, the noised stream conditions on the clean stream's state from all earlier blocks and on itself. Setting the diffusion block equal to the recurrence kernel's 64-token chunk makes this nearly free: the chunked kernel already materializes the state entering each chunk, so the clean state at the start of block $n$ costs nothing extra to expose, and the noised stream needs no scan of its own. In the noised pass, the kept attention layers attend to clean keys from previous blocks and noised keys within the block. The first position of each block (the \emph{anchor}) is always fed clean. Per-block noise levels are drawn uniformly on $[0.2, 1]$ with the $1/t$ loss weighting of \citet{sahoo2024mdlm}, and the output is shifted by one position so that the output head keeps the teacher's next-token semantics.

\textbf{Budget.} Pretraining runs at sequence length 512 and $8192$ tokens per step, for $85{,}449$ steps (700M tokens drawn from a 2.1B-token code-and-text corpus) on one TPU v5e-8 host, at learning rate $3{\times}10^{-4}$ up to step $50{,}729$ and $2{\times}10^{-4}$ after it. A $4{,}000$-step supervised phase follows: chat-formatted examples at sequence length 512 and batch 16, with only the response region noised (mask ratio drawn from $U[0.3,1.0]$) and ten percent of each batch drawn from the pretraining corpus as a regularizer.

\textbf{Negative result 1: a representation-alignment auxiliary does nothing.} We added a term aligning the student's clean-stream hidden states to the frozen teacher's at each layer boundary, expecting it to preserve teacher structure through the objective change. Measurement showed otherwise. Cosine alignment was already ${\approx}94\%$ saturated at initialization, because the copied projections and frozen backbone anchor the hidden states on their own. The term did not improve autoregressive retention, inflated logit scale roughly twofold, and cost $10$--$15\%$ of step time. We disabled it.

\textbf{Negative result 2: the learning-rate ceiling appears late.} At learning rate $3{\times}10^{-4}$ the run diverged twice, both times deep into training (around steps $50$--$52$k and $56$k). The first forced a rollback to step $20{,}430$ and was not logged in enough detail to diagnose. The second, whose per-step log was captured, showed escalating oscillation in loss over several hundred steps, then a spike, a partial recovery and a collapse that drove accuracy to ${\approx}0.02$. No single step crossed our per-step loss-spike threshold, so a step-level monitor registered nothing. After a rollback to step $50{,}729$ and a reduction to $2{\times}10^{-4}$, training was stable for the remaining ${\approx}35$k steps, as $2{\times}10^{-4}$ had been for the $1.5$B pilot conversion that preceded this one. The failure is invisible to the obvious monitor and appears only after tens of thousands of stable steps.

\section{Discussion}\label{sec:discussion}

\textbf{What conversion costs.} Teacher knowledge survives the conversion only partially and only in-distribution (\S\ref{sec:stage1}); the in-sequence binding faculty does not survive at all. It is absent rather than degraded, and diffusion continual pretraining does not bring it back. It has to be retrained deliberately, and the companion paper's mechanism account \citep{anatomyrecall} says why: the deficit is a hole in what training covered, so only training that covers it can close it.

\textbf{Closed-loop curricula are the transferable lesson.} An open-loop curriculum is a clock; a closed-loop one is a controller; the difference is $1/3$ versus $3/3$ on identical seeds. The same distinction predicts the companion bench's length boundary, where an open-loop schedule collapses to $0/6$ at sequence length 512 \citep{anatomyrecall}. Wherever a curriculum has a difficulty knob and a measurable competence signal, gating the former on the latter converts a lottery into a method. The natural extension, and the one all of our open negatives point to, is to gate the training \emph{budget} the same way: a fixed step count is itself an open-loop decision about a widely varying random variable.

\textbf{What the coverage boundary means.} A converted model whose retrieval was trained over a narrow symbol pool scores $0.000$ outside it, and that boundary survived pool randomization, three gate policies and a scale change. Curriculum-installed retrieval is a circuit over the embeddings that took part in retrieval training, which makes token coverage a first-class dimension of the training distribution.

\textbf{Two monitoring lessons from the block-diffusion conversion.} Both of its negative results have the shape of the open-loop failures elsewhere in this paper: a decision fixed in advance, and a cheap measurement that would have overturned it, taken only after the cost was paid. A per-step monitor saw nothing while the run diverged over hundreds of steps; a representation-alignment term that looked principled at design time turned out, once measured, to be saturated already at initialization.

\section{Limitations}\label{sec:limits}

\begin{enumerate}[leftmargin=1.4em,itemsep=2pt]
\item \textbf{Token generality is unsolved.} Every solved model reads $0.000$ on held-out token bands (\S\ref{sec:coverage}). We make no claim that the curriculum installs a general retrieval faculty; the annealing fix that our own diagnosis motivates is untested.
\item \textbf{Language-modeling transfer is narrow.} Stage 1 distilled on a 12M-character slice of one corpus. At $8$B, held-out agreement is $0.544$ and off-domain perplexity retention is near zero (\S\ref{sec:stage1}), so the converted students are not general-purpose language models.
\item \textbf{Three seeds per setting.} Conversion-scale rows rest on three seeds (one for the wikitext arm). $3/3$, $2/3$ and $1/1$ are small-$n$ rates, and no pairwise contrast in this paper is significance-tested.
\item \textbf{Fixed budgets confound several negatives.} The $8$B seed-2 miss, the RWKV-7 no-onset arm and the pool-annealing arm are all consistent with budget truncation of a late onset rather than method failure. Until the closed-loop budget is run, those three results remain open.
\item \textbf{One probe family.} Retrieval is measured with a single multi-query recall probe whose format is close to that of the training episodes. The evaluation uses held-out bindings and a held-out token band, but the format is still close to training, and we do not claim broad downstream retrieval.
\item \textbf{The block-diffusion conversion is only partly evaluated.} It has a measured held-out NLL and a ${\approx}4.1\times$ perplexity ratio against the teacher, and its free-running generation collapses into repetition; systematic generation, retrieval-transfer and instruction-following evaluation are deferred to a forthcoming results paper.
\end{enumerate}

\section*{Reproducibility Statement}
Per-arm training telemetry (gate cap, retrieval EMA, avalanche onset), evaluation outputs for every arm including the failed ones, and checkpoints for most arms (not the unseeded pilot) are retained in object storage. The companion bench's code and results are public at \url{https://github.com/JIBSIL/dualgoose}; the conversion code reported here is not yet released.

\section*{Acknowledgments}
This work used computing resources provided by the Rosen Center for Advanced Computing (RCAC) at Purdue University~\citep{McCartney2014}.

\section*{Impact Statement}
This work concerns making pretrained language models cheaper to re-architect, which reduces the compute needed to deploy capable models and, with it, the barrier to deploying them without scrutiny. The retrieval boundary we report is also safety-relevant: a converted model can look competent on its training distribution and show no retrieval just outside it, a failure mode that perplexity-based evaluation does not surface.

\setlength{\bibsep}{1pt}

\end{document}